%% file: main.tex
\documentclass[a4paper, 10pt, conference]{ieeeconf} 

\IEEEoverridecommandlockouts                              
\usepackage{graphics}
\usepackage{epsfig}
\usepackage{times}
\usepackage{amsmath}
\usepackage{float}
\usepackage{svg}
\usepackage{subfigure}
\usepackage{sidecap}
\sidecaptionvpos{figure}{c}
\usepackage[font=footnotesize]{caption}
\title{\LARGE \bf
Controlled Face Manipulation and Synthesis for Data Augmentation
}

\author{\parbox{16cm}{\centering
    {\large Joris Kirchner$^{1,2,}${\Large $^\star$}, 
    Amogh Gudi$^{1,2,}${\Large $^\star$}, 
    Marian Bittner$^{1,2}$, 
    Chirag Raman$^2$}\\
    {\normalsize
    $^1$ Vicarious Perception Technologies (VicarVision)\\
    $^2$ Delft University of Technology\\
    {\Large $^\star$} Equal contribution.
    }}
}
\usepackage{url}
\usepackage{hyperref}
\usepackage{amsfonts, mathtools}
\let\labelindent\relax
\usepackage{enumitem}
\usepackage{fancyhdr}

\begin{document}

\maketitle

\begin{abstract}
\input{sections/abstract}
\end{abstract}

\section{INTRODUCTION}
\input{sections/introduction}

\section{RELATED WORK}
\input{sections/related_work}

\section{METHOD}
\input{sections/method}

\section{EXPERIMENTS AND RESULTS}
\input{sections/experiments}

\section{DISCUSSION}
\input{sections/discussion}

{\small
\bibliographystyle{ieeetr}
\bibliography{references}
}

\onecolumn
\appendix
\input{supplementary}

\end{document}

%% file: sections/abstract.tex
Deep learning vision models excel with abundant supervision, but many applications face label scarcity and class imbalance.
Controllable image editing can augment scarce labeled data, yet edits often introduce artifacts and entangle non-target attributes.
We study this in facial expression analysis, targeting Action Unit (AU) manipulation where annotation is costly and AU co-activation drives entanglement.
We present a facial manipulation method that operates in the semantic latent space of a pre-trained face generator (Diffusion Autoencoder).
Using lightweight linear models, we reduce entanglement of semantic features via (i) dependency-aware conditioning that accounts for AU co-activation, and (ii) orthogonal projection that removes nuisance attribute directions (e.g., glasses), together with an expression neutralization step to enable absolute AU edit.
We use these edits to balance AU occurrence by editing labeled faces and to diversify identities/demographics via controlled synthesis.
Augmenting AU detector training with the generated data improves accuracy and yields more disentangled predictions with fewer co-activation shortcuts, outperforming alternative data-efficient training strategies and suggesting improvements similar to what would require substantially more labeled data in our learning-curve analysis.
Compared to prior methods, our edits are stronger, produce fewer artifacts, and preserve identity better.

%% file: sections/introduction.tex
Getting finely-labeled image data can be slow and expensive, and the distribution of collected data suffers from frequency biases (e.g., under-representation of certain ethnicities \cite{buolamwini2018gender}). 
An alternative is to create additional examples by editing existing images along the label axes (e.g., changing facial expression intensity \cite{InterFaceGAN, GANSpace}). 
This, however, suffers from the risk that an edit can unintentionally change other attributes (e.g., identity, lighting, another expression), producing noisy labels \cite{bau2019semantic}. 
What remains unsolved is how to achieve targeted edits that modify only the intended attribute while leaving all other factors unchanged.

Among existing methods for image manipulation, generative adversarial networks (GAN) based editors require a separate, often computationally intensive and error-prone encoding method \cite{e4e, DAE}; while text-to-image diffusion (e.g., ControlNet \cite{ControlNET}) does not offer sufficiently precise control for disentangled editing of specific attributes \cite{MagicFace}. 
Diffusion Autoencoders (DiffAE) \cite{DAE} pair an encoder with a diffusion decoder, exposing a semantic latent space which allows for precise image reconstruction from embeddings and where edits can be expressed as simple directions.
However, making complex edits remains challenging, especially in the domain of face manipulation where expressions simultaneously affect different parts of the face.
In this work, we propose a technique for obtaining fine-grained attribute control with a generic, pre-trained generator, without retraining a large generative model from scratch.
Specifically, we build on DiffAE~\cite{DAE}, which provides an encoder--decoder interface and a semantic latent space suitable for such manipulation.

\begin{figure}
    \centering
    \includegraphics[trim={0 4.7cm 4.29cm 0},clip,width=1\linewidth]{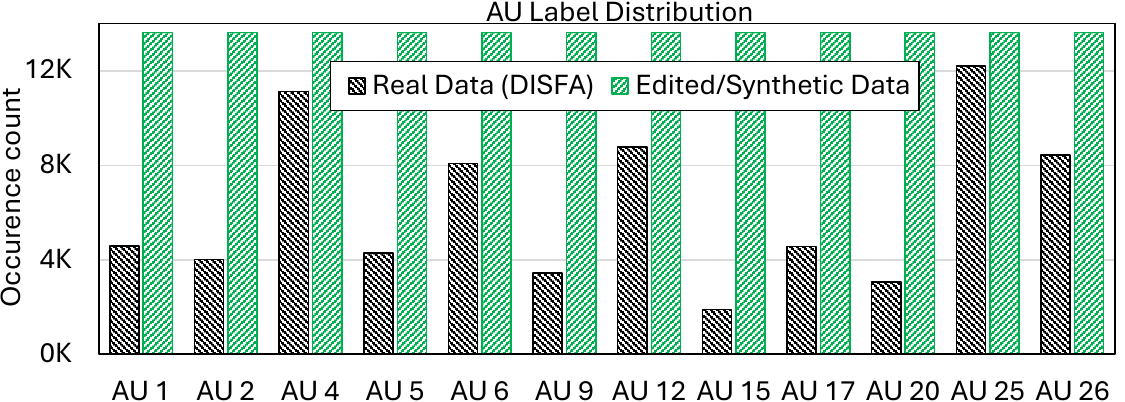}
    \caption{Distribution of AU occurrence labels in datasets. Naturally occurring distribution of AU labels in real data (DISFA \cite{DISFA, DISFA+}) is highly skewed (black). In contrast, our method for controlled editing/synthesis allows for generation of datasets with a balanced distribution (green).}
    \label{fig:distribution}
\end{figure}

In facial expression analysis, the Facial Action Coding System (FACS) decomposes expressions into Action Units (AUs) \cite{ekman1978facial}. 
AUs often co-occur in natural facial expressions, producing statistical entanglements: AU activations correlate with each other (e.g., AU1 vs. AU2) and nuisance attributes (e.g., pose, eyeglasses, lighting). Editing specific AUs with models capturing natural variations also affects non-target attributes like other AUs, identity, and pose; creating artifacts and label noise.
Moreover, facial AU datasets are class-imbalanced and long-tailed, where rare AUs are under-represented \cite{DISFA} (see Fig.~\ref{fig:distribution}) and demographics are unevenly covered.
Manual AU labeling requires specialized experts (certified FACS coders \cite{facswebsite}), making large-scale annotation costly.
Consequently, contemporary image augmentation techniques fail to address these deficits that limit training generalization. 

We target precisely this challenging domain of expression editing to demonstrate AU-level control in facial images that preserves other attributes.
Specifically, we propose a technique that leverages  
(i) dependency-aware conditioning that blocks unwanted AU co-activation when estimating edit directions, and 
(ii) orthogonal projection to remove components aligned with nuisance or competing attributes on the generator's encoding. 
We also apply an expression neutralization step that suppresses any pre-existing AU activations before applying edits, enabling absolute AU edits.
We demonstrate our method for augmenting supervised training data for AU detection via (a) editing existing labeled faces to balance AU occurrences, and (b) synthesizing new faces with targeted demographic attributes (e.g., gender, age) and AUs; 
which results in improvement in detection accuracy and disentanglement.

Concretely, we make the following contributions:
\begin{itemize}[leftmargin=*]
\item We introduce a framework to re-purpose a generic, pre-trained face image generator (DiffAE) into an AU-controllable editor/synthesizer using lightweight models in the generator's semantic latent space, avoiding task-specific generator retraining.
\item We propose two effective methods to reduce entanglement in edits: (i) dependency-aware conditioning to suppress unwanted AU co-activations; (ii) latent-space projection of generator embeddings to remove entangled/nuisance attribute directions.
\item We present a procedure to sample new identities from the generator, neutralize pre-existing facial expressions, and impose desired AU configurations while fixing other attributes; enabling control over identity and demographic distribution in generated faces.
\item We empirically show better single/multi-AU editing accuracy and identity preservation robustness compared to other methods; 
and demonstrate that augmenting supervised training with our generated data gives better results than alternate data-efficient training strategies.
\end{itemize}

%% file: sections/related_work.tex
\textbf{Obtaining latent representations.}
Rather than making changes directly in the pixel space \cite{deepdream}, an effective approach to making semantic edits in an image is to first encode it into a latent space and then edit the code before decoding \cite{SeFa, WarpedGANSpace}.
Latent spaces from pre-trained GANs enable such encodings for face manipulation \cite{StyleGAN, BIGGAN, ProgGAN, StyleGAN2}, and direction discovery in those spaces supports attribute control \cite{GANSpace, InterFaceGAN, SeFa, WarpedGANSpace, nada}.
Because these generators are not trained as encoders, inversion methods are employed to obtain codes for real images, with e4e~\cite{e4e} providing a practical encoder-based solution. 
In practice, inversion remains imperfect and can shift facial details, including Action Units, which motivates an alternative encoder–decoder with stronger reconstruction fidelity for downstream AU editing; we explore this in our work.

\textbf{Diffusion-based encoders and DiffAE.}
Diffusion probabilistic models provide high-fidelity generative priors and stable likelihood-based training \cite{DDPM, DDIM}, and Diffusion Autoencoders (DiffAE) expose a semantic latent space that supports direct manipulation while maintaining reconstruction quality \cite{DAE}. 
DiffAE separates a stochastic code (fine details e.g., hair pattern, skin texture) from a semantic code (`global' attributes e.g., facial expressions, age, gender), allowing edits to be applied primarily in the semantic space \cite{DAE}. 
This structure aligns with our goal of precise AU control with minimal identity or background drift; hence, we use DiffAE's latent representation to showcase our model.

\textbf{Latent-space manipulation: directions and paths.}
Attribute editing can be achieved by moving codes along latent directions that induce semantic changes~\cite{GANSpace, InterFaceGAN, SeFa, StyleFlow, StyleRig, WarpedGANSpace, nada}.
Supervised approaches estimate directions using labeled attributes (e.g., linear separators for target factors) \cite{InterFaceGAN, StyleFlow}, while unsupervised approaches discover axes by analyzing internal latent structure or feature responses \cite{GANSpace, SeFa, WarpedGANSpace, StyleRig}.
These mechanisms provide efficient, low-compute controls that are compatible with encoder--decoder pipelines.
In our work, we adopt supervised linear directions in DiffAE’s semantic space to target specific AUs.

\textbf{Text-to-image editing for faces.}
Text-conditioned manipulation models, enable broad, instruction-like edits, which could be used for AU editing~\cite{nada, diffusionclip}. 
However, generic text-to-image pipelines prioritize versatility over precise, disentangled control of fine-grained facial components such as individual AUs, and may entangle pose, lighting, or accessories with expressions during editing. 
Our setting instead requires targeted, label-faithful AU manipulation; we therefore choose latent-space edits in an autoencoding framework over general text-driven editing. 

\textbf{AU editing methods.}
Expression editing with AU-level control has been addressed with GAN-based and diffusion-based pipelines. 
Early AU manipulation demonstrated feasibility but introduced artifacts in challenging regions \cite{GANimation, GANimation_article}. 
Subsequent methods increase fidelity by leveraging modern generators and architectural priors, including StyleGAN-based pipelines with content/exemplar guidance \cite{StyleAU, StyleGAN} and dual-branch designs that aim to separate AU intensity from other facial factors \cite{AUEditNet, StyleGAN2}. 
Recent diffusion-based approaches improve realism and introduce AU conditioning via external detectors and attribute controllers for pose/background consistency \cite{MagicFace}. 
Our approach differs by allowing usage of any face generator without using an attribute controller or requiring an AU detector, focusing instead on aligned faces and linear edits; this design yields controllable, disentangled AU changes while keeping the overall method relatively simple.

%% file: sections/method.tex
\begin{figure*}
    \centering
    \includegraphics[width=.9\linewidth]{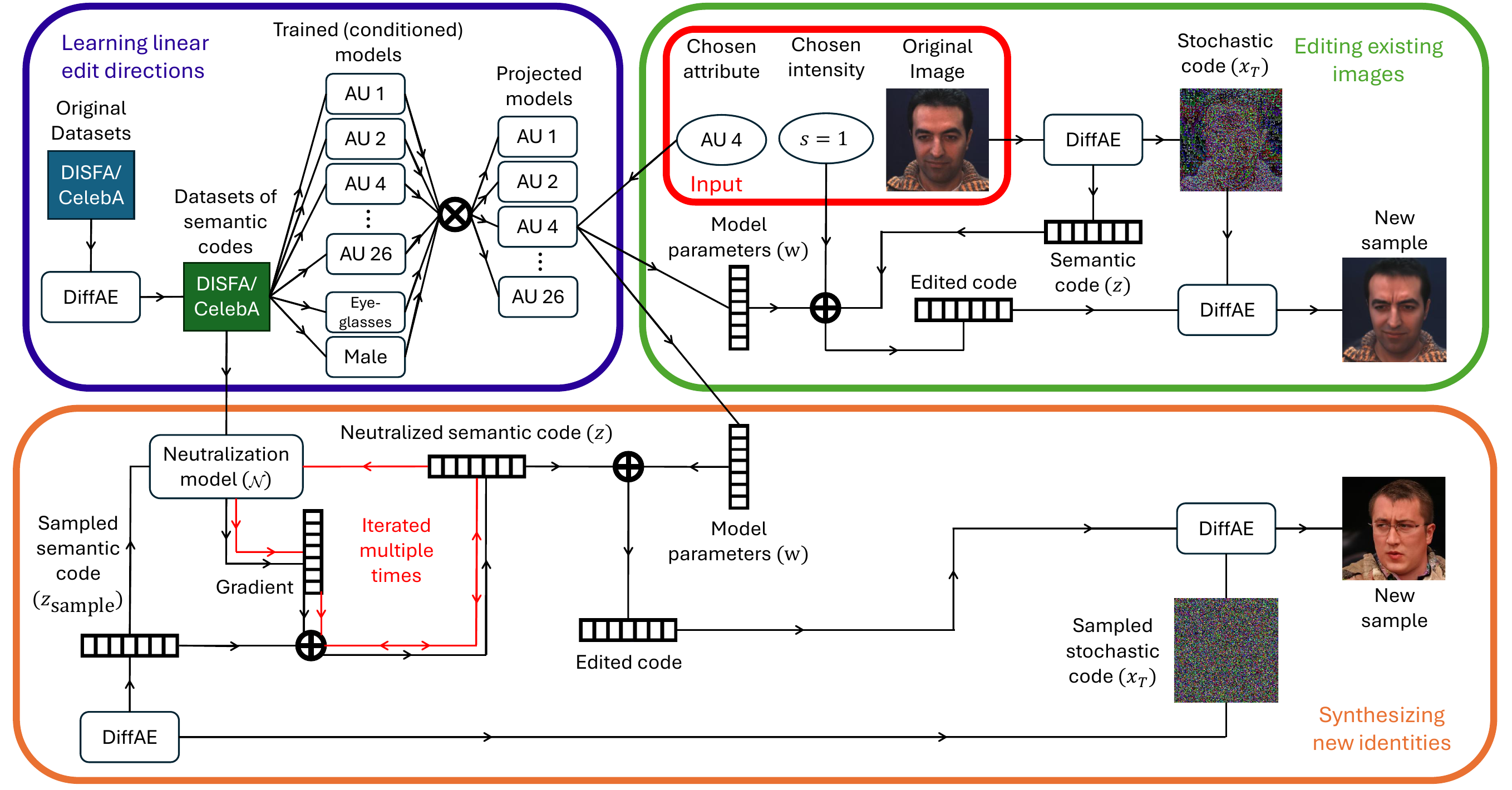}
    \caption{
    Overview of our method. (Top-Left) Learning linear edit directions on semantic codes, where AU models are conditioned on other AUs. Afterwards, AU directions are projected on other possible nuisance attributes. (Top-Right) \emph{Editing existing images:} encode $(x_T,z)$, pick target AU, obtain disentangled direction $w$, set $z\leftarrow z+s\,w$, decode with original $x_T$. (Bottom) \emph{Synthesizing new faces:} sample $(x_T,z_{\text{sample}})$ from DiffAE, optionally accept by demographic predictors, neutralize $z_{\text{sample}}$ via the neutralization model $\mathcal{N}$ by optimizing Equation~\eqref{eq:optimization_objective}, then edit and decode.
    }
    \label{fig:model_schematic}
    \vspace{-0.25cm}
\end{figure*}

Our pipeline operates in a face generator's latent space to (i) learn linear edit directions for AU control, while suppressing entanglements, (ii) edit existing neutral images to obtain new samples with distinct AU activations, and (iii) synthesize new demographically balanced identities by neutralizing random samples to a zero-AU state. Fig.~\ref{fig:model_schematic} summarizes these components and their data flow. To explain the method, we assume DiffAE is used. However, the method generalizes to be used in combination with other face generators that allow for encoding and decoding.

\subsection{Learning linear directions to edit existing images}
\textbf{Semantic directions for continuous AU control.}
This step turns linear predictors on DiffAE semantics into continuous controllers for AU intensities. 
For an encoding $(x_T, z)$, where $x_T$ is the stochastic code and $z$ is the semantic code, we follow the procedure proposed in \cite{DAE} to only edit $z$. 

Linear classifiers/regressors in latent space (e.g., logistic regression or support vector machines (SVM) as in \cite{DAE,InterFaceGAN}) are commonly defined for binary attributes, whereas AU labels are continuous on $[0,1]$. Let a linear predictor be $\hat y=z w + w_0$ for semantic code $z$, parameters $(w,w_0)$, and attribute score $\hat y$. We use the parameter vector $w$ itself as an edit direction: moving $z \leftarrow z + s\,w$ monotonically increases the predictor output. For SVMs, this aligns with the normal of the decision boundary (used in \cite{InterFaceGAN}); for logistic regression, the input-gradient is a positive scalar multiple of $w$ because the sigmoid derivative is positive, preserving direction.  

\begin{figure}
    \centering
    \includegraphics[width=.8\linewidth]{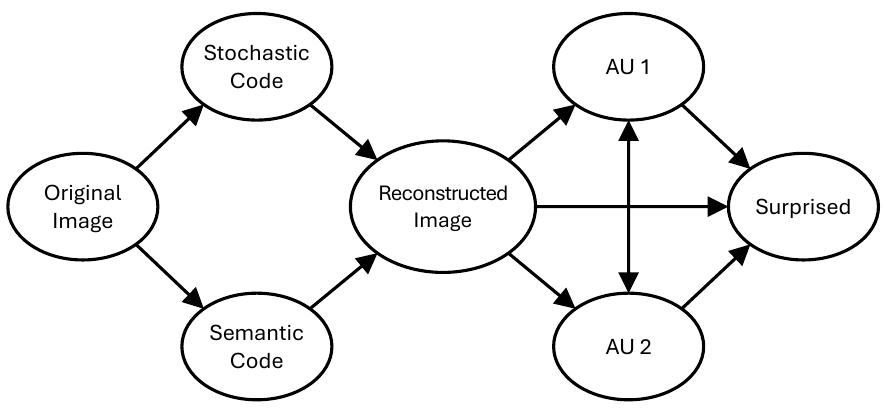}
    \caption{An illustrative DAG showing an example of AU relations.
    The intensity of AU 1 is defined by the reconstructed image, but also by AU 2 as they are often activated together. 
    Conditioning on a correlated AU blocks leakage from that AU to the target AU. 
    At the same time attributes that are influenced by AU1, called colliders (e.g., \emph{surprised}), should not be conditioned upon to avoid opening spurious paths.
     }
    \label{fig:causal_dag}
\end{figure}

\textbf{Disentanglement controls in latent space.}
The aim of this step is to reduce leakage between target AUs with other AUs, as well as with nuisance attributes.  
We apply two complementary techniques:
 \begin{enumerate}
    \item \emph{Conditional dependence}: When labels for potentially entangled AUs are available, the target predictor is conditioned on the other AU(s), blocking paths that drive co-activation. We reason about paths via a directed acyclic graph (DAG) (for example, Fig.~\ref{fig:causal_dag}); we assume labels transfer from original to reconstruction and that the stochastic code has negligible direct effect on AU predictions, avoiding costly conditioning on $x_T$. Models that edit a specific AU (e.g. AU1) are trained with the labels of other AUs (e.g. AU2) to block learning of backdoor paths. Care is taken with colliders (e.g., \emph{surprised} in Fig.~\ref{fig:causal_dag}) to avoid introducing spurious dependence.
    \item \emph{Orthogonal projection}: When labels for a nuisance factor (e.g., eyeglasses) are missing, the target direction is projected onto the orthogonal complement of a set of nuisance directions as in \cite{InterFaceGAN}. Projection can weaken the target edit if attributes genuinely overlap. We therefore choose projection sets empirically.
\end{enumerate}

\subsection{Synthesizing new faces}
\textbf{Controlled sampling for demographic balance.}
Employing the procedure outlined in \cite{DAE}, this step balances demographic attributes like gender and age in synthetic identities.
Linear attribute predictors are trained on semantic codes for coarse demographics (e.g. age, gender). Conditional sampling proceeds via acceptance–rejection \cite{DAE}: draw many $z$, retain only those for which the predictors indicate the desired demographic bins.
This simple conditional sampling in semantic space provides balanced demographic groups without altering the core editing pipeline or requiring additional generative training.

\textbf{Neutralizing randomly sampled faces.}
With this step, we produce labeled synthetic faces by first driving the sampled faces to a neutral expression state with all AUs inactive (neutralization).
This detector-guided neutralization supplies zero-AU starting points for controlled single-AU edits on sampled identities. This procedure is critical as the sampled identities have unknown AU activations and relative AU edits do not allow for correct labeling of images.

This step has two parts: (i) training the neutralization model $\mathcal{N}$ once before synthesis; and (ii) using $\mathcal{N}$ to neutralize a sampled semantic code $z_{\text{sample}}$ by minimizing the differentiable loss $\mathcal{L}(\mathcal{N}(z_{\text{sample}}),\mathbf{y^\star})$ toward a neutral target $\mathbf{y^\star}$ (with unchanged nuisance attributes), while keeping the detector weights frozen. Note that this procedure would also work for non-neutral targets, thereby removing the need for subsequent linear edits. However, the advantage of the neutralization model is that it only needs to neutralize, making its training easier and more stable.

To optimize $\mathcal{N}$ for neutralization, it is important to achieve high AU recall. We compute recall by binarizing continuous AU predictions at an activation threshold (e.g., $0.1$) to emphasize AU presence/absence. High recall minimizes false negatives, ensuring active AUs are detected with higher sensitivity and thus reliably suppressed during neutralization.

\textbf{Optimization objective and step policy.}
Here, we detail the loss and update strategy used during neutralization.  
The objective combines an attribute loss and a proximity regularizer,
\begin{equation}\label{eq:optimization_objective}
\min_{z}\;\mathcal{L}(\mathcal{N}(z),\mathbf{y}^\star)\;+\;\lambda\|z-z_{\text{sample}}\|_2^2.
\end{equation}
Rather than a single large update, multiple small steps are taken with gradient re-evaluation to reduce overshoot and preserve realism, similar to the procedure in \cite{deepdream} for pixel space manipulation. This allows the usage of Adam \cite{Adam} to optimize $z$, while freezing detector parameters.
To prevent drift off the data manifold, the latent proximity term keeps edits close to the initial code.

%% file: sections/experiments.tex
\subsection{Implementation details}

\textbf{Datasets.}
We conduct all experiments on the combination of DISFA~\cite{DISFA} and DISFA+~\cite{DISFA+} (collectively called \emph{DISFA}), which together contain approximately 45K posed and spontaneous images from 32 subjects intensity-annotated for 12 AUs (1, 2, 4, 5, 6, 9, 12, 15, 17, 20, 25, 26).
The label distribution is visualized in Fig.~\ref{fig:distribution}.
We adopt an 80:20 subject-exclusive split into training and validation sets.
In addition, we employ the AU-labeled FEAFA~\cite{yan2019feafa} dataset (intensity labels for AUs 2, 4, 9, 12, 17, 20, 26) and BP4D~\cite{zhang2014bp4d} dataset (presence labels for AUs 1, 2, 4, 6, 9, 12, 15, 17, 20) for cross-dataset validation.

We use the CelebA~\cite{CelebA} dataset for removing nuisance entanglements by learning specific semantic attribute directions: eyeglasses, beard, gender, and age group. Finally, to study the effect of unsupervised pretraining for the AU detector, we use the unlabeled FFHQ dataset~\cite{StyleGAN}; this is the same dataset used to train the DiffAE generator~\cite{DAE}.

\textbf{Data augmentation.}
We employ two strategies for data augmentation in our experiments:
\begin{enumerate}
    \item \emph{Editing real faces.} For every neutral face image in the DISFA training split (i.e., no active AUs), we create 12 edited variants by activating exactly one AU per image. This procedure yields approximately 160K images and results in a balanced AU distribution (see Fig.~\ref{fig:distribution}).
    \item \emph{Synthetic face generation.} We sample neutral identities from DiffAE while maintaining balanced demographics (equal male/female counts and equal representation across three age groups: young, middle, old). Each neutral face is then edited into 12 variants with one AU activated per image, producing another $\approx$160K images so that the synthetic set matches the edited set in size.
\end{enumerate}
In the remainder of this section, we refer to the union of these two sets as \emph{generated} data; their individual contribution is analyzed later in the ablation studies (Sec.~\ref{sec:ablations}).

\noindent
\textbf{Hyperparameters.}

\emph{Generator.}
We employ the DiffAE~\cite{DAE} face generation model pretrained on FFHQ~\cite{StyleGAN}.
We use a combination of support vector~\cite{vapnik1997support} and ridge regression~\cite{hilt1977ridge} models (default hyperparams, $\alpha=10$) for AU editing, and use a neural network for neutralization (architecture in Appendix).

\emph{Neutralization.}
The neutralization model is optimized using a mean squared error loss for AU prediction and a binary cross-entropy loss for binary attributes. Training is terminated if the total AU recall does not improve for 15 epochs.
For neutralization, a dropout rate of 0.2 is applied to the latent code, with a regularization parameter $\lambda = 0.004$. Optimization is stopped when the 50-step moving average of the objective fails to decrease by more than 0.002 over 150 steps. To avoid sigmoid saturation, logits are used for binary attributes, while keeping sigmoids during model training.

\emph{AU detection.}
For downstream AU detection, we train a \textit{MobileNetV3-Small}~\cite{howard2019searching} backbone CNN to predict AU intensities (as a 12-dimensional vector, one per AU) from 224$\times$224 RGB face crop inputs.
The network is randomly initialized (unless stated otherwise) and trained using the Adam optimizer~\cite{Adam} (default hyperparameters) with early stopping based on validation loss (for 40 epochs).
All training runs are repeated five times and the results are aggregated.

\emph{Other methods.}
MagicFace~\cite{MagicFace} and NNCLR~\cite{dwibedi2021little} were implemented using default hyperparameters. StyleGAN-NADA~\cite{nada} uses small edit guidelines, enabling {\small\verb|improve_shape|} with 20 training iterations.

\subsection{Synthesis/editing of facial expressions}
\label{sec:editing}

\begin{figure*}
    \centering
    \includegraphics[trim={0 13.3cm 0 0cm},clip,width=\linewidth]{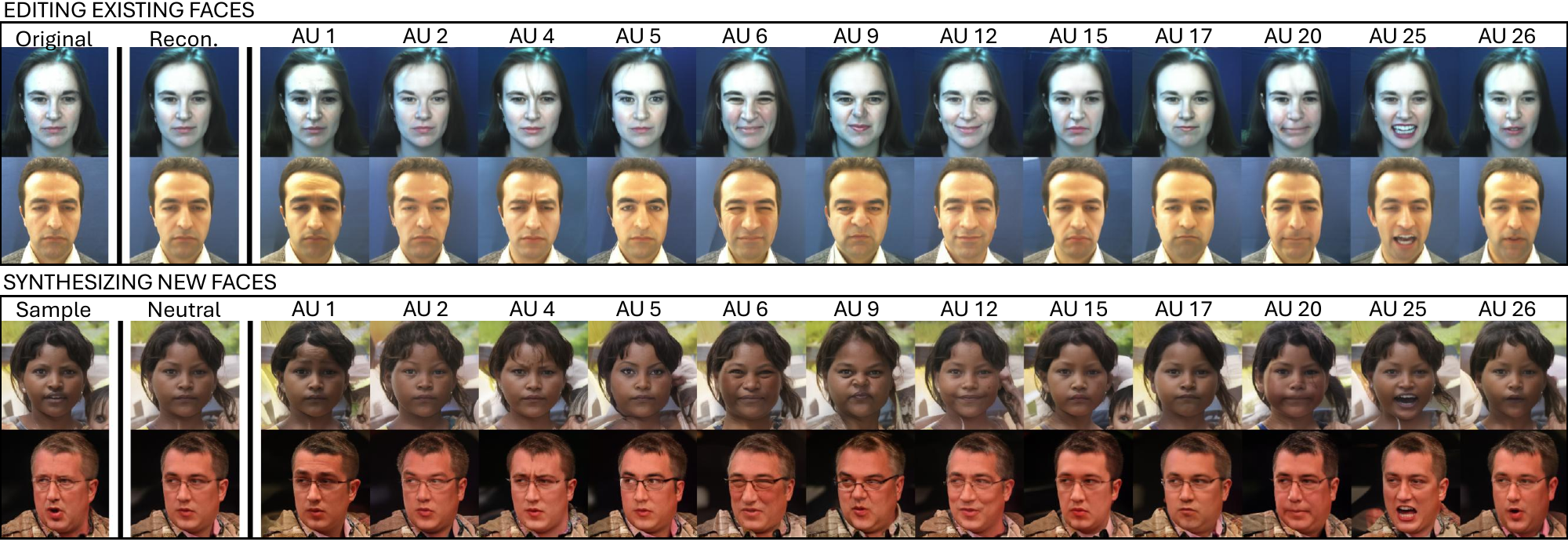}
    \caption{Examples of controlled AU editing (top half) and controlled synthesis of new faces (bottom half).
    Editing examples (top) uses existing neutral images from the DISFA dataset \cite{DISFA}, where reconstruction is obtained by encoding and decoding without editing. 
    Synthesis examples (bottom) contains random identities sampled from DiffAE containing arbitrary facial expressions. which is first neutralized by the non-linear procedure to suppress pre-existing expressions. 
    Thereafter, each AU is modified by $+1$ using the proposed method, corresponding to FACS activation level E.
    Editing of images is largely without major artifacts, and neutralization succeeds in deactivating AUs; but complex expressions may leave small residuals, and editing may slightly alter background of synthetic images.}
    \label{fig:random_example}
    \vspace{-0.25cm}
\end{figure*}

\textbf{Qualitative AU editing.}
Fig.~\ref{fig:random_example} shows examples of controlled AU editing on real DISFA images (top) and on randomly sampled DiffAE identities (bottom).
Overall, the edits are localized and visually plausible across different identities and demographics, and typically preserve global factors such as pose, illumination, and background.
In challenging cases, edits may slightly affect stochastically encoded details (e.g., small background changes for some synthesized samples).

\textbf{Neutralization enables absolute edits.}
A key step in our pipeline is \emph{neutralization}: before applying a target AU configuration, we suppress any pre-existing expression in the semantic code using the neutralization network.
This is vital because applying a linear edit direction to an already expressive face is otherwise \emph{relative} (i.e., it modifies an unknown baseline), leading to inconsistent outcomes across identities and source frames.
As shown in Fig.~\ref{fig:random_example} (bottom), neutralization typically removes existing expression well, enabling \emph{absolute} edits with predictable intensity changes.
Residual activations may remain for complex expressions (e.g., slightly open mouth in the last row), but outputs are substantially closer to neutral than the originals.

\begin{figure*}
    \centering
    \includegraphics[trim={0 0.2cm 0 0},clip,width=\linewidth]{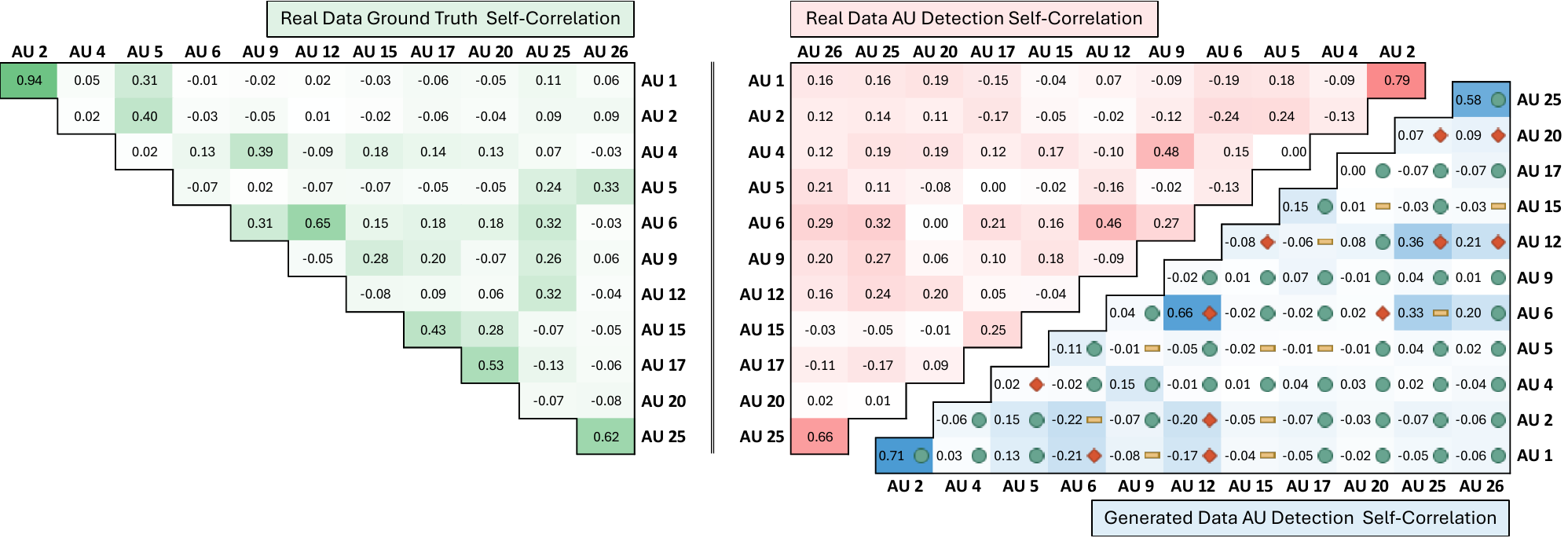}
    \caption{
    (Right) Comparison of correlation between detected AUs in the real DISFA data (top/red) and the edited/synthetic data (bottom/blue), as estimated by an AU detection tool (FaceReader~\cite{facereader}) as a stand-in for human annotators. 
    The color heatmap represents the amplitude of the correlation, while the green/yellow/red markers in the edited/synthetic data table denote the intensity of change w.r.t. correlations in the real data (green = reduction, red = increase, yellow = no significant change). 
    Detected AUs in the edited/synthetic data are much less correlated with each other on average (0.09) as compared to the detected AUs in the real data (0.16).
    (Left) For reference, the inter-AU correlations between the ground truth labels in DISFA are also shown. These exhibit similar patterns as the detector's AU estimates on the real images, suggesting limited influence of detector bias.
    }
    \label{fig:correlation_matrix}
\end{figure*}

\textbf{Disentanglement analysis via inter-AU correlation.}
We quantify how much our generation reduces spurious AU co-activation by comparing inter-AU correlations in real vs generated data.
We estimate AU intensities using an AU detection tool (FaceReader~\cite{facereader}, selected after an empirical comparison against other alternatives~\cite{baltruvsaitis2016openface, ertugrul2019afar} where it showed higher accuracy) as a proxy for human annotators, and compute pairwise correlations as shown in Fig.~\ref{fig:correlation_matrix} (right).
Many AU pairs are moderately to strongly correlated in real DISFA images, whereas the generated set shows substantially weaker dependencies.
The DISFA \emph{ground-truth} correlation structure (Fig.~\ref{fig:correlation_matrix}, left) closely matches the detector’s correlations on real DISFA, suggesting these dependencies are largely intrinsic to the dataset rather than an artifact of the detector.
Overall, generated images reduce inter-AU correlations from 0.16 to 0.09 (average absolute), indicating that controllable generation lowers typical co-activation biases.

\subsection{Training with generated data augmentation}
\label{sec:aug_results}

\begin{figure}
    \centering
    \includegraphics[trim={0 7.8cm 5.7cm 0},clip, width=\linewidth]{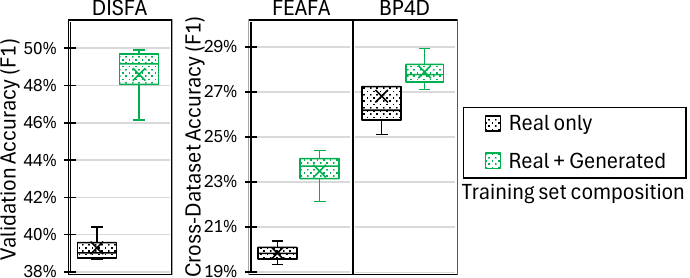}
    \caption{
    Box plots showing AU detection accuracy (F1 scores, averaged over AUs) for models trained with (Real + Generated) and without (Real only) generated augmentation on DISFA~\cite{DISFA,DISFA+} (left), as well as cross-dataset evaluation on FEAFA~\cite{yan2019feafa} and BP4D~\cite{zhang2014bp4d} (right). 
    Generated augmentation consistently improves F1 on all three benchmarks.
    }
    \label{fig:au_detection_f1}
\end{figure}

\textbf{Improvement in AU detection across datasets.}
We evaluate how generated augmentation affects AU detection accuracy (Fig.~\ref{fig:au_detection_f1}).
On DISFA, it improves mean F1 from $\approx$39\% to $\approx$49\%, equivalent to a 25\% improvement over baseline.
We also observe consistent cross-dataset gains on FEAFA (+20\%) and BP4D (+4\%), with the smaller BP4D improvement possibly due to differing labeling standards since it was not labeled for AU intensities (only presence).
These results suggest that the benefit of generated augmentation transfers beyond in-distribution validation.

\begin{SCfigure}
    \centering
    \includegraphics[trim={0 5.35cm 4.75cm 0},clip,width=0.6\linewidth]{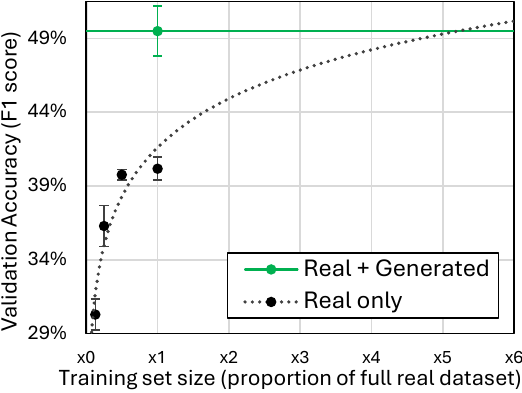}
    \caption{
    AU detection accuracy (F1) of baseline models trained on varying amounts of real data (learning curve). The horizontal line shows performance with generated augmentation; error bars denote standard deviation over five runs. A POW3 fit~\cite{viering2022shape} suggests $\approx5\times$ more real data is needed to match the augmented accuracy.
    }
    \label{fig:learning_curve}
\end{SCfigure}

\textbf{Learning curve analysis.}
To contextualize the accuracy improvement on DISFA, we repeat the baseline training (with no generated augmentation) while progressively sub-sampling the real DISFA training set.
Fig.~\ref{fig:learning_curve} plots the resulting learning curve and overlays the performance of generated augmentation training as a horizontal reference.
Extrapolating the baseline learning curve (power law with offset~\cite{viering2022shape}) suggests that substantially more real labeled data would be required to match the performance achieved by generated augmentation (in our experiments, $\approx$5$\times$ more).

\begin{SCfigure*}
    \centering
    \includegraphics[trim={0 3.7cm 3.2cm 0},clip,width=0.55\textwidth,page=1]{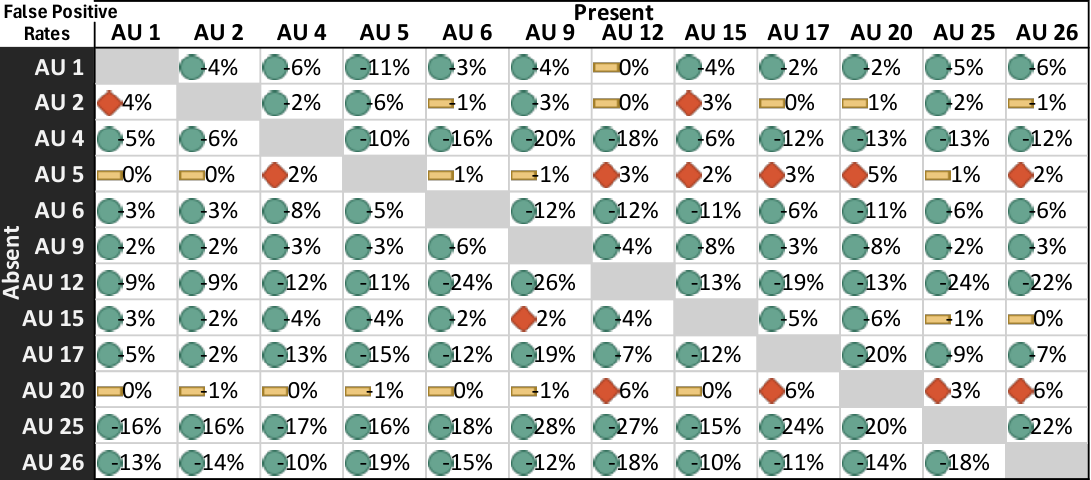}
    \caption{
    Difference in validation set false positive rates (FPRs) for AU pairs (where one AU is absent while another is present) between models trained with and without our generated augmentation technique.
    The green, yellow, and red markers represent the intensity of change (green = reduction, red = increase, yellow = no significant change).
    These false positives are a sign of the model over-exploiting the entanglements/correlations between AUs to make predictions.
    Training with our generated augmentation results in disentangled learning: yielding a lower false positive rate for a majority of AUs as compared to training with real data alone, with an average reduction of 7.4\% points.
    }
    \label{fig:fprate_table}
\end{SCfigure*}

\textbf{Reduced cross-AU false positives.}
We examine whether training on disentangled data changes the trained AU detector’s error structure.
If a model exploits spurious AU correlations (e.g., predicting AU6 whenever AU12 is present), it will produce false positives when one AU is absent but the other is present.
We therefore compare AU-pair false positive rates (FPR) with and without generated augmentation, where AU-pair FPR is the rate of falsely detecting an absent AU conditioned on another AU being present.
As shown in Fig.~\ref{fig:fprate_table}, generated augmentation reduces AU-pair false positives for most entangled pairs and lowers the average AU-pair FPR by 7.4 percentage points.
This suggests more independent AU detectors with reduced reliance on co-activation shortcuts.

\subsection{Comparison with other methods}
\subsubsection{Facial expression manipulation}\hfill
\vspace{0.1cm}

\begin{figure*}
    \centering
    \includegraphics[trim={0 15.7cm 2.55cm 0},clip,width=\linewidth]{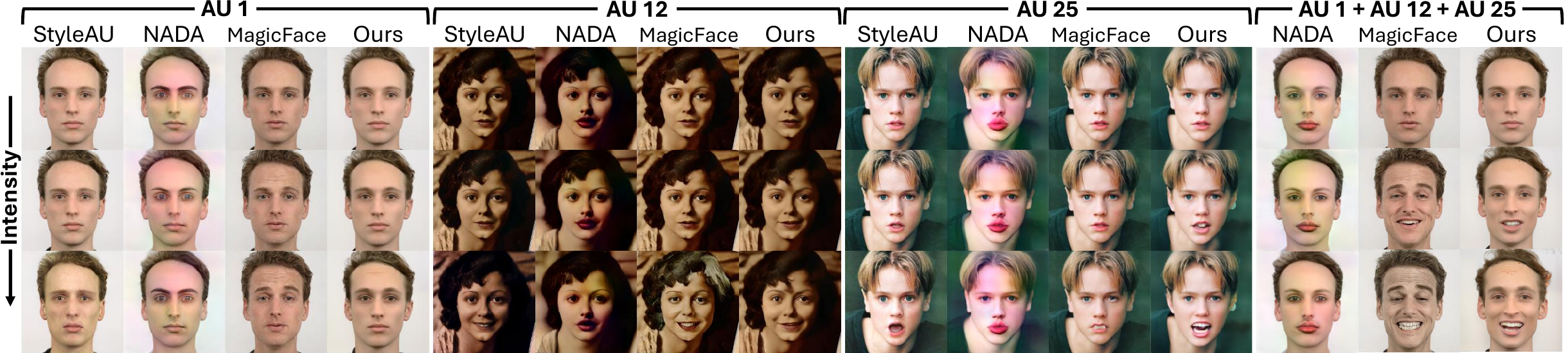}
    \caption{Qualitative comparison of facial expression editing from our method with StyleAU~\cite{StyleAU}, StyleGAN-NADA~\cite{nada} and MagicFace~\cite{MagicFace} (image sourced from \cite{StyleAU}). Editing is done at various intensity levels (increasing from top to bottom) loosely corresponding to FACS AU intensity levels from A to E:
    StyleAU $\{1,3,5\}$, StyleGAN-NADA $\{0,0.4,0.8\}$, MagicFace $\{0,4,6\}$, and our method $\{0,0.5,1\}$.
    For StyleGAN-NADA, the images are generated with target class: ``\textit{face with [AU description] increased by [intensity]}".
    Our method retains the AUs in the original face better, while also often performing a strong plausible edit and introducing fewer artifacts under single and multi-AU editing.
    }
    \label{fig:SOTA_comparison}
    \vspace{-0.25cm}
\end{figure*}

\textbf{Stronger visual edits with fewer artifacts.}
Fig.~\ref{fig:SOTA_comparison} visually compares our edits with other methods: StyleAU~\cite{StyleAU}, StyleGAN-NADA~\cite{nada}, and MagicFace~\cite{MagicFace}.
At higher edit strengths, these methods exhibit artifacts (e.g., extreme wrinkles, color distortion), whereas our results remain clean while producing plausibly stronger and localized AU activation.

\begin{figure}
    \centering
    \includegraphics[trim={0cm 7.5cm 6.4cm 0cm},clip,width=\linewidth, page=1]{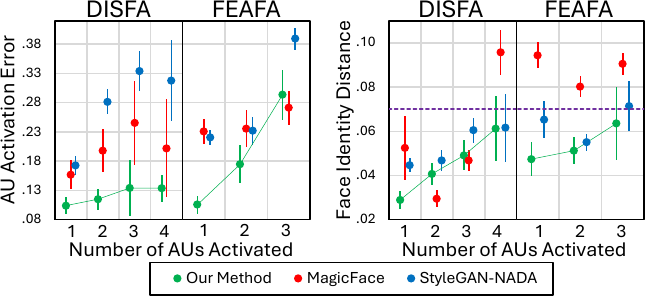}
    \caption{
    Comparison of AU-editing fidelity (left) and identity preservation (right) across manipulation methods as a function of the number of simultaneously edited AUs.
    Left: MAE between AU activations estimated on the edited image and on an identity-matched real target image exhibiting the desired AU enactment. 
    Right: identity drift measured as the cosine similarity distance between face recognition embeddings of the edited image and the corresponding real target image (same identity), with the purple line indicating the face recognition threshold.
    Legend: Error bars denote standard error.
    Across all AU counts, our method achieves the lowest AU error, indicating the closest match to target AU configurations, while maintaining low identity drift.
    As expected, identity drift increases with the number of edited AUs, but our edits remain within the recognition threshold.
    }
    \label{fig:other_edit_methods}
    \vspace{-0.25cm}
\end{figure}

\textbf{Higher AU-editing accuracy.}
We next quantify how accurately different editing methods can match a target AU configuration (StyleAU~\cite{StyleAU} not evaluated here due to lack of publicly available code).
For a set of identity-matched image pairs (source, target) from DISFA and FEAFA datasets, where the target exhibits the desired AU activations, we edit the source image towards the target AU configuration using each method.
We then compute the mean absolute error (MAE) between an AU detector's (FaceReader~\cite{facereader}) estimated AU intensities on the edited image and on the target image.
As shown in Fig.~\ref{fig:other_edit_methods} (left), our method achieves lower MAE than MagicFace and StyleGAN-NADA across all settings, indicating a closer match to the intended AU configuration.
MAE increases for all methods as more AUs are edited simultaneously, reflecting the greater difficulty of multi-AU control.
On DISFA, the gap to baselines widens with AU count, suggesting stronger robustness under multi-AU edits.
On FEAFA, the largest gap is seen for one AU, and performance degrades sharply with higher AU counts.

\textbf{Better identity preservation.}
To quantify identity retention under editing, we measure identity drift using a face recognition model~\cite{dlibfacereco, dlib09}, computed as cosine distance between embeddings of the original and edited images from DISFA and FEAFA.
Fig.~\ref{fig:other_edit_methods} (right) shows that identity drift increases for all methods as more AUs are edited, reflecting the challenges of multi-AU editing.
Nevertheless, our method stays below the identity-matching threshold (dashed line) across all AU counts on both datasets.
StyleGAN-NADA~\cite{nada} induces only minor visual changes in this setup, yielding low distances; whereas MagicFace~\cite{MagicFace} shows substantial identity drift (over the threshold), especially at higher AU counts.

\vspace{0.1cm}
\subsubsection{Alternative data-efficient training strategies}\hfill
\vspace{0.1cm}

We compare our generated augmentation with two common data-efficient training approaches: (i) inverse-frequency loss reweighting to counter class imbalance~\cite{zhao2025loss}, and (ii) unsupervised pretraining to mitigate limited labels~\cite{erhan2010does, dwibedi2021little}.
Here, \emph{reweighting} weights each AU by the inverse of its training set frequency, and \emph{pretraining} applies NNCLR~\cite{dwibedi2021little} to the AU detector backbone on FFHQ~\cite{StyleGAN} before supervised AU training.
FFHQ dataset is used for a fair comparison since DiffAE was trained on it.

\begin{figure}
    \centering
    \includegraphics[trim={0 6.7cm 3.85cm 0},clip, width=\linewidth]{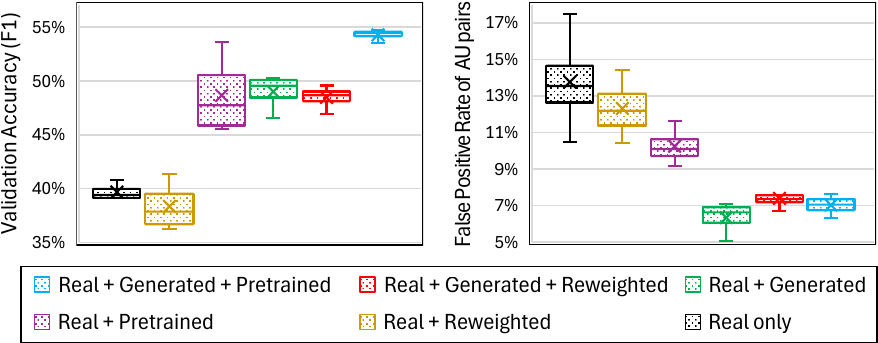}
    \caption{Comparison of AU-detection accuracies (averaged F1 scores over AUs; top) and false positive rates of AU pairs (where one AU is absent while another is present; bottom) of models trained using different data-efficient training setups. 
    Combining generated augmentation with reweighing does not yield further improvement in accuracy, but combining with pretraining yields the highest accuracy. 
    While reweighing and pretraining increasingly lower the AU-pair FPR, the largest improvement is caused by the introduction of our generated augmentation.
    }
    \label{fig:other_methods}
    \vspace{-0.5cm}
\end{figure}

\textbf{Highest AU detection accuracy.}
Fig.~\ref{fig:other_methods} (left) shows that reweighting alone does not improve over the baseline, whereas generated augmentation yields substantially higher accuracy.
Combining reweighting with generated augmentation provides no additional gain, suggesting that augmentation already mitigates imbalance inherently.
Pretraining also improves accuracy and is complementary: pretraining followed by generated augmentation achieves the best performance, significantly outperforming either component alone while using the same unlabeled and labeled datasets.

\textbf{Lowest cross-AU false positives.}
Since F1 does not capture reliance on AU co-activation, we compare average AU-pair false positive rates (FPR) across strategies (as described in Sec.~\ref{sec:aug_results}), see Fig.~\ref{fig:other_methods} (right).
The baseline has the highest AU-pair FPR ($\approx$14\%), reduced by reweighting ($\approx$12\%) and further by pretraining ($\approx$10\%).
Generated augmentation yields a much lower AU-pair FPR ($\approx$6.5\%), and combining it with pretraining or reweighting does not further change FPR.
Notably, even when F1 is similar (Fig.~\ref{fig:other_methods}, left), generated augmentation achieves markedly lower AU-pair FPR, suggesting less overfitting on inter-AU correlations.

\subsection{Ablation studies}
\label{sec:ablations}

\subsubsection{Disentanglement controls}\hfill
\vspace{0.1cm}

\begin{figure}
    \centering
    \includegraphics[trim={0cm 4.75cm 1.1cm 0cm},clip,width=\linewidth]{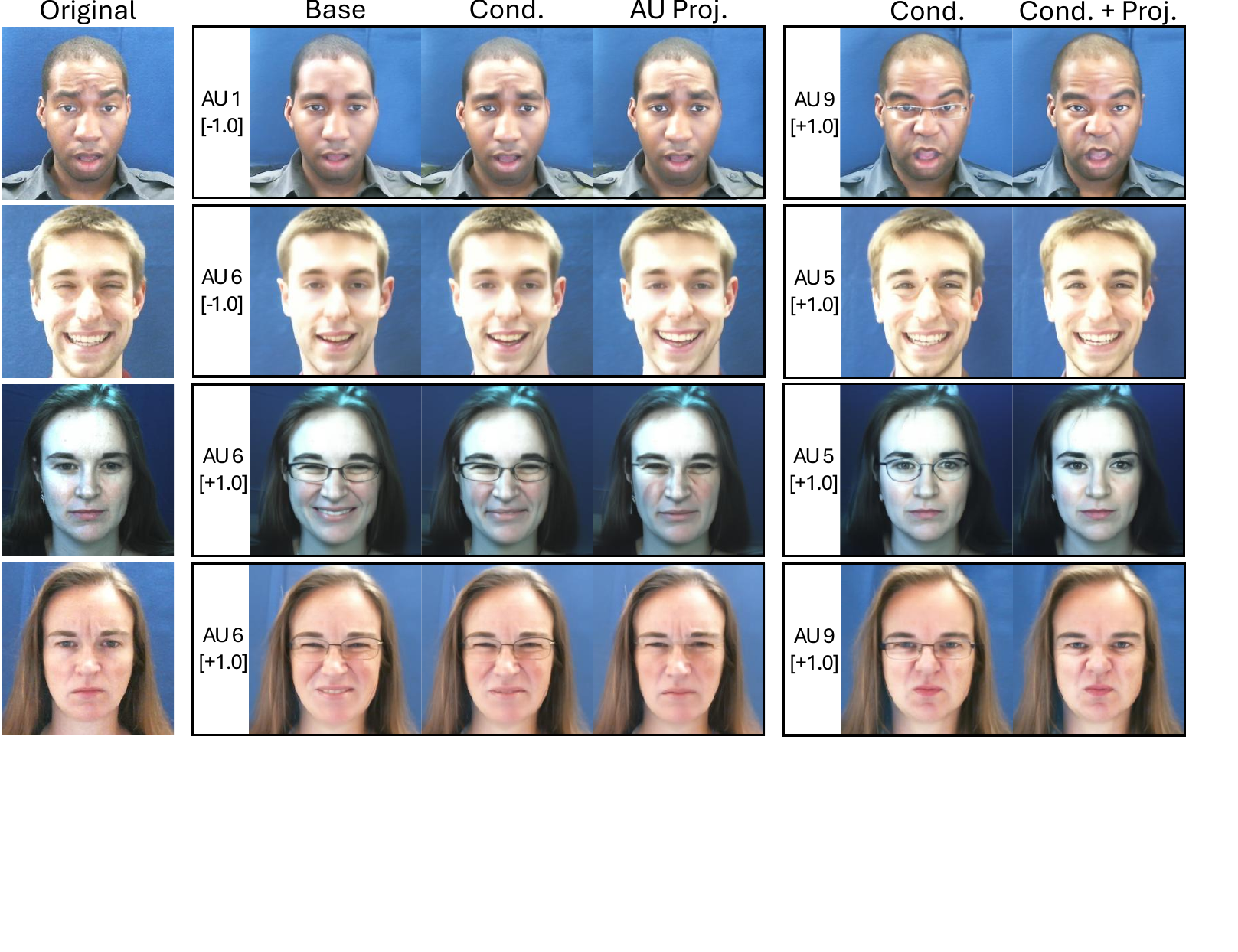}
    \caption{Qualitative comparison showcasing the effect of conditioning (Cond.) vs
    projecting on other AUs (AU Proj.) and the effect of projecting conditioned models on other entangled attributes (Cond.+Proj.). Brackets under the edited AU names denote the strengths and directions of the edits: $\pm$1 approximately corresponds to FACS intensity level E. Conditioning or projection reduces entanglement with other non-target AUs (e.g., AU1--AU2, AU6--AU12). Conditioning tends to preserve mild, plausible overlapping effect, while projection removes them entirely. Furthermore, projecting on nuisance attributes like glasses removes entanglement with them.
    }
    \label{fig:au_disentanglement}
\end{figure}

\textbf{Qualitative.}
Fig.~\ref{fig:au_disentanglement} shows that naive linear editing (``Base'') can inadvertently change correlated AUs.
Conditioning reduces predictable AU leakage by accounting for co-activation during direction estimation, while projection more aggressively removes residual overlap by eliminating components aligned with competing/nuisance directions.
For instance, Base edits often co-modify AU1 and AU2; conditioning better isolates the target AU.
At higher intensities, coupling between AU6 and AU12 is reduced by projecting the AU6 direction against AU12, whereas conditioning may retain a small, physiologically plausible overlap.
Fig.~\ref{fig:au_disentanglement} further shows that projection against an ``eyeglasses'' direction (learned from CelebA) suppresses spurious glasses artifacts that can appear when editing AU5/AU9.

\begin{table}[]
    \centering
    \caption{
    Effect of conditioning and projection on fine-grained edit fidelity, measured by SSIM w.r.t real target image (mean $\pm$ standard error). 
    Conditioning and projection progressively improve SSIM over the baseline.
    }
    \begin{tabular}{c|ccc}
        &Baseline & Conditioning & Cond. + Projection\\\hline
         SSIM & $0.8125\pm 0.0033$ & $0.8233\pm 0.0033$ & $\mathbf{0.8300\pm 0.0032}$
    \end{tabular}
    \label{tab:ssim_ablation}
    \vspace{-0.5cm}
\end{table}

\textbf{Quantitative.}
We additionally quantify how closely the edited result matches a real target image of the same identity exhibiting the intended AU activation.
Tab.~\ref{tab:ssim_ablation} reports the aggregated structured similarity (SSIM)~\cite{wang2004imageSSIM} between edited images and their corresponding real target images.
The results show that conditioning improves SSIM over the baseline editor and adding projection further increases SSIM, which aligns with the qualitative observations.

\vspace{0.1cm}
\subsubsection{Ablation over augmentation strategies}\hfill
\vspace{0.1cm}

\begin{figure}
    \centering
    \includegraphics[trim={0cm 7.5cm 0.35cm 0cm},clip,width=\linewidth,page=1]{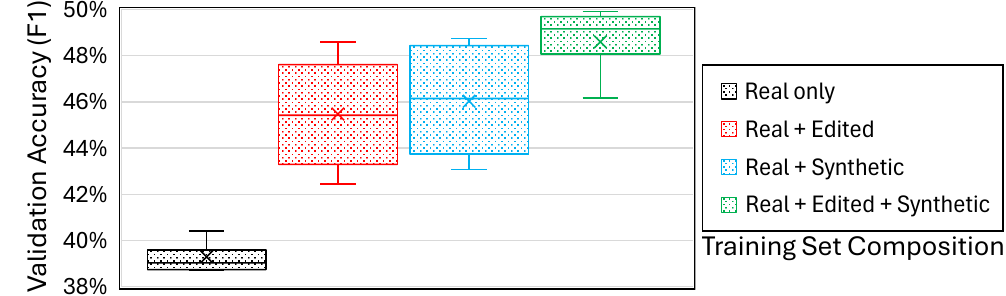}
    \caption{
    Box-plot comparison of the AU detection accuracy (F1 score) of models trained using different data augmentation strategies.
    The best performance is achieved from augmenting with both edited and synthetic images, while augmenting only with either one of them yields a comparatively lower accuracy (that is still higher than using real data only).
    }
    \label{fig:aug_ablation}
\end{figure}

Finally, we isolate the contributions of the two augmentation strategies used to form our generated data.
Fig.~\ref{fig:aug_ablation} compares training on real images, augmenting with edited faces, augmenting with synthetic faces, and their combination.
Both strategies yield substantial and similar improvements individually, while their combination performs best, suggesting complementary benefits: editing could be adding expression diversity in realistic settings, whereas synthesis adds identity/demographic diversity.

%% file: sections/discussion.tex
\noindent
\textbf{Summary.} We presented a practical framework for converting a generic, pretrained face generator into an \emph{AU-controllable} face editing tool.
Moreover, downstream AU detection trained on data augmented using our technique improves performance, while also reducing inter-AU entanglement. These findings suggest that controllable semantic-space editing is a viable path to data augmentation in settings where AU labels are expensive and long-tailed, and where reducing reliance on co-activation shortcuts is as important as improving overall performance.

Our controllable augmentation addresses two challenges of natural data: long-tailed AU distribution and strong AU co-activation.
Our generated data exhibits substantially lower inter-AU correlations than real images (Fig.~\ref{fig:correlation_matrix}), and training with our data augmentation yields higher accuracy (Fig.~\ref{fig:au_detection_f1}) and reduces cross-AU false positives for many entangled AU pairs (Fig.~\ref{fig:fprate_table}).

Beyond downstream AU detection, we compare editing quality to alternative manipulation methods.
Qualitatively, our edits remain localized and visually clean at higher edit strengths (Fig.~\ref{fig:SOTA_comparison}).
Quantitatively, our approach achieves lower MAE of AU estimates between edited and real images for both single- and multi-AU editing (Fig.~\ref{fig:other_edit_methods}, left), and preserves identity better across all AU counts (Fig.~\ref{fig:other_edit_methods}, right).

Finally, we compared our generated augmentation to two common data-efficient training strategies: inverse-frequency loss reweighting and self-supervised pretraining (NNCLR~\cite{dwibedi2021little}).
Reweighting alone does not improve average F1, and adding it to generated augmentation yields no further gains (Fig.~\ref{fig:other_methods}, left); which is consistent with the fact that our augmentation explicitly balances AU occurrences by construction.
Pretraining provides a strong improvement and is complementary: combining it with generated augmentation achieves the best accuracy.

\subsection{Limitations and Future Work}
\begin{itemize}[leftmargin=*]
    \item Our method assumes access to a suitable pre-trained face generator. While this avoids training from scratch, it may limit adaptation where such a generator is unavailable.
    \item Although we support multi-AU editing, our augmentation balances individual AUs. Targeting under-represented \emph{AU combinations} could yield a richer training distribution.
\end{itemize}

\subsection{Ethical considerations}
Our work uses only established public face datasets and demonstrates synthesis/editing of faces to improve facial expression analysis.
However, controllable face synthesis may be misused (e.g., identity manipulation or deceptive content generation), so future data releases should include safeguards such as restricted access and intended-use statements.
Additionally, our use of external AU detection tools and face recognition models for some analyses raises bias concerns; we mitigate this by referencing AU ground-truth labels as a partial sanity check, and encourage future work to validate findings with human coders when possible.

%% file: supplementary.tex
\begin{figure}[h]
    \centering
    \includegraphics[width=\linewidth]{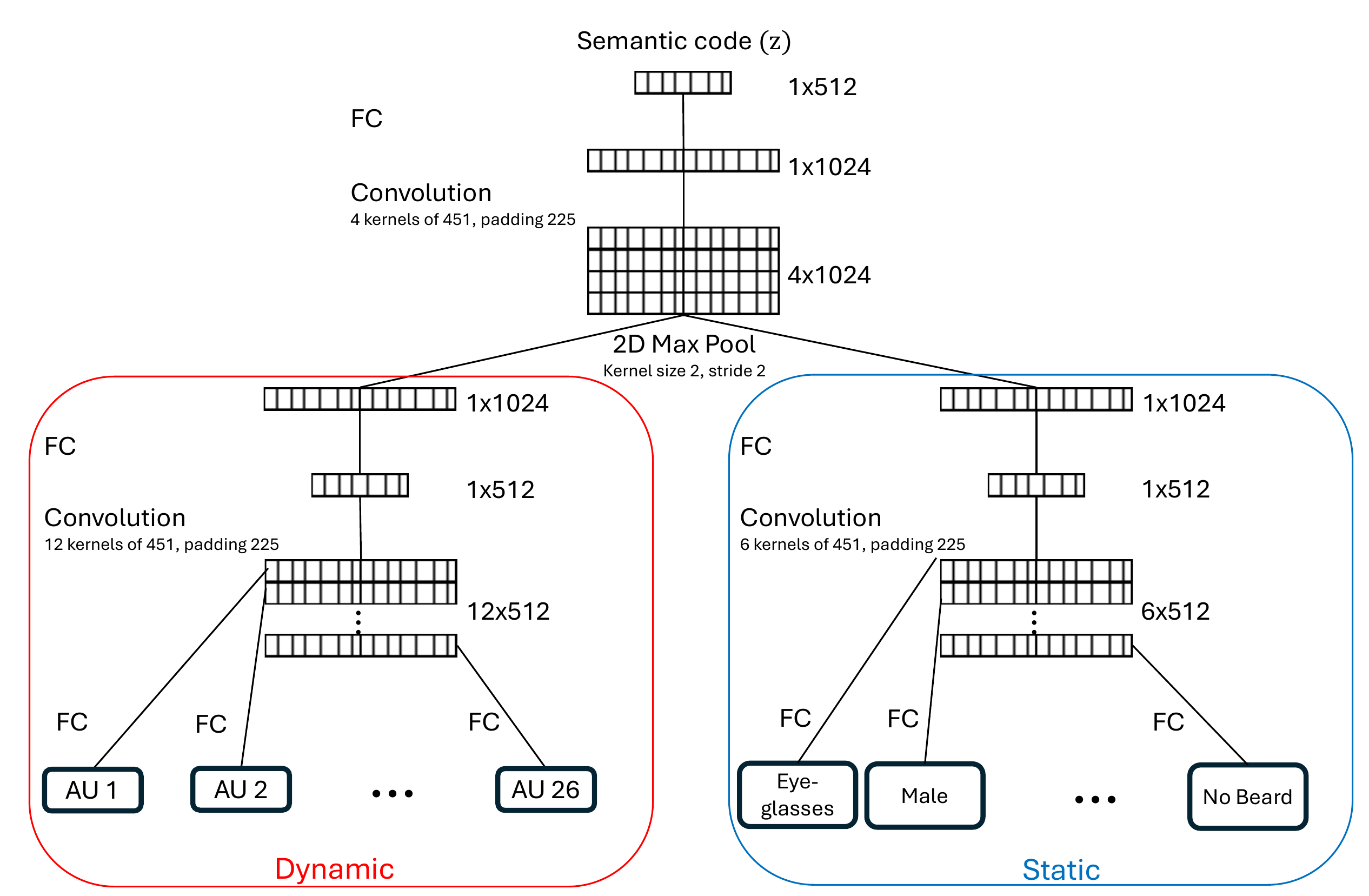}
    \vspace{0.1cm}
    \caption{Neural network architecture of the neutralization model $\mathcal{N}$. Fully Connected layers are abbreviated with FC. After each linear layer, the SiLU~\cite{SiLU} activation function is applied. The network splits into two branches. A dynamic branch that leads to AU predictions, and a static part that leads to the prediction of the other static features. Finally, each attribute has a FC layer of 512, only used for predicting that specific attribute.}
    \label{fig:nn_architecture}
\end{figure}

\vspace{0.75cm}
Fig.~\ref{fig:nn_architecture} shows the neural network architecture of the neutralization model $\mathcal{N}$ used for neutralization.

